\documentclass[format=acmsmall, review=false, screen=true]{acmart}
\usepackage{epsfig}
\usepackage{graphicx}
\usepackage{amsmath}

\usepackage{multirow}
\usepackage{mathrsfs}
\usepackage{enumitem}
\usepackage{makecell}
\usepackage{tabulary}
\usepackage{pifont}
\usepackage{subfigure}
\usepackage{bm}
\begin{document}

\title[Uni-EDEN: Universal Encoder-Decoder Network by Multi-Granular Vision-Language Pre-training]{Uni-EDEN: Universal Encoder-Decoder Network by\\ Multi-Granular Vision-Language Pre-training}

\author{Yehao~Li}
\affiliation{%
  \institution{JD AI Research}
  \city{Beijing}
  \country{China}}
\email{yehaoli.sysu@gmail.com}

\author{Jiahao~Fan}
\affiliation{%
  \institution{Shanghai Jiao Tong University}
  \city{Shanghai}
  \country{China}}
\email{jiahaofan@sjtu.edu.cn}

\author{Yingwei~Pan}
\affiliation{%
  \institution{JD AI Research}
  \city{Beijing}
  \country{China}}
\email{panyw.ustc@gmail.com}

\author{Ting~Yao}
\authornote{Corresponding author.}
\affiliation{%
  \institution{JD AI Research}
  \city{Beijing}
  \country{China}}
\email{tingyao.ustc@gmail.com}

\author{Weiyao~Lin}
\affiliation{%
  \institution{Shanghai Jiao Tong University}
  \city{Shanghai}
  \country{China}}
\email{wylin@sjtu.edu.cn}

\author{Tao~Mei}
\affiliation{%
  \institution{JD AI Research}
  \city{Beijing}
  \country{China}}
\email{tmei@jd.com}


\begin{abstract}
Vision-language pre-training has been an emerging and fast-developing research topic, which transfers multi-modal knowledge from rich-resource pre-training task to limited-resource downstream tasks.
Unlike existing works that predominantly learn a single generic encoder, we present a pre-trainable Universal Encoder-DEcoder Network (Uni-EDEN) to facilitate both vision-language perception (e.g., visual question answering) and generation (e.g., image captioning). Uni-EDEN is a two-stream Transformer based structure, consisting of three modules: object and sentence encoders that separately learns the representations of each modality, and sentence decoder that enables both multi-modal reasoning and sentence generation via inter-modal interaction. Considering that the linguistic representations of each image can span different granularities in this hierarchy including, from simple to comprehensive, individual label, a phrase, and a natural sentence, we pre-train Uni-EDEN through multi-granular vision-language proxy tasks: Masked Object Classification (MOC), Masked Region Phrase Generation (MRPG), Image-Sentence Matching (ISM), and Masked Sentence Generation (MSG). In this way, Uni-EDEN is endowed with the power of both multi-modal representation extraction and language modeling. Extensive experiments demonstrate the compelling generalizability of Uni-EDEN by fine-tuning it to four vision-language perception and generation downstream tasks.

\end{abstract}

\setcopyright{acmcopyright}
\acmJournal{TOMM}
\acmYear{2021} \acmVolume{1} \acmNumber{1} \acmArticle{1} \acmMonth{1} \acmPrice{15.00}\acmDOI{10.1145/3473140}

%
%

\ccsdesc[300]{Computing methodologies~Scene understanding}
\ccsdesc[500]{Computing methodologies~Computer vision representations}
\ccsdesc[300]{Computing methodologies~Natural language generation}
\ccsdesc[300]{Information systems~Multimedia content creation}
\ccsdesc[300]{Computing methodologies~Transfer learning}

\keywords{Vision-language pre-training, Encoder-decoder networks}


\maketitle

\renewcommand{\shortauthors}{Y. Li et al.}
\section{Introduction}

\begin{figure*}[!tb]
\vspace{-0.0in}
\centering {\includegraphics[width=0.96\textwidth]{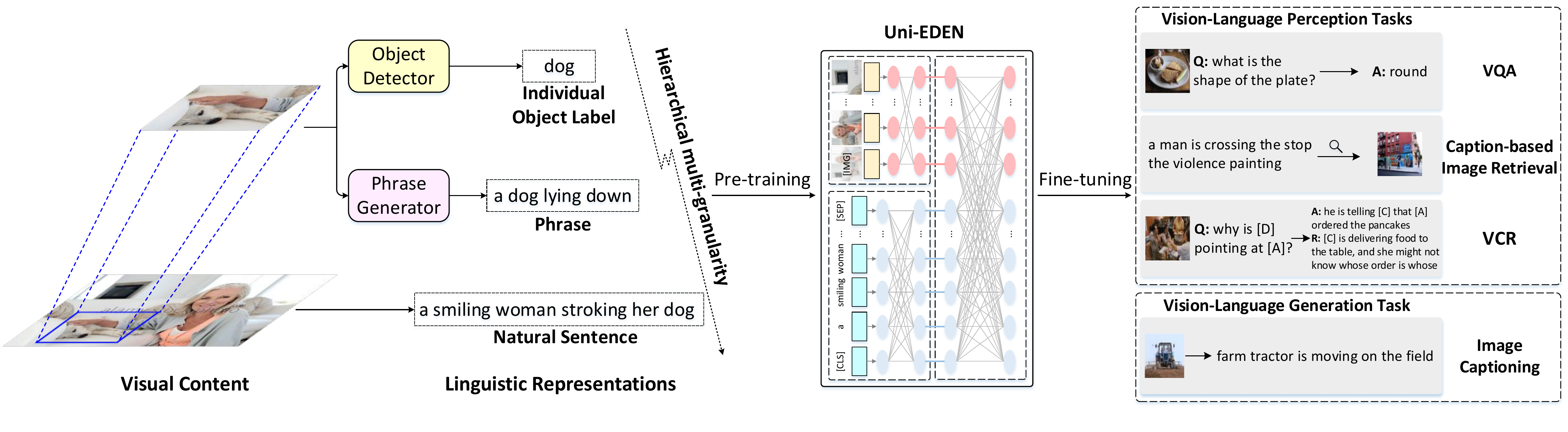}}
\vspace{-0.26in}
\caption{\small The linguistic representations of visual content can span different granularities. For example, by applying object detector over an image, each detected region can be represented as an individual object label. Going beyond the individual label, we can utilize phrase generator to produce a more comprehensive phrase to depict that region. Furthermore, the holistic image can be described with a natural sentence that recapitulates more details. In this work, we exploit such hierarchical multi-granularity to pre-train a universal encoder-decoder network (Uni-EDEN). The pre-trained Uni-EDEN benefits both vision-language perception and generation downstream tasks.}
\label{fig:figintro}
\vspace{-0.32in}
\end{figure*}

Vision and language are two fundamental capabilities of human intelligence. The interactions in between support the uniquely human capacity, such as answering open-ended questions w.r.t the given image (visual question answering \citep{antol2015vqa,gao2019dynamic,kim2018bilinear}), generating multimodal responses based on user intentions (multimodal task-oriented dialog \cite{nie2019multimodal}), or describing what they see with a natural sentence (image captioning \citep{anderson2017bottom,yao2018exploring,cornia2018paying,wang2018image,yang2019image,vinyals2015show}). With the development of deep learning techniques, there has been a steady momentum of breakthroughs that push the limits of vision-language tasks \cite{wang2020visual,qu2020context}. Despite having promising quantitative results, the achievements rely heavily on the requirement of large quantities of task-specific annotations (e.g., image-question-answer triplets/image-sentence pairs) for such neural model learning. This severely hinders the scalability and generalization of vision-language techniques when only limited annotations are available. To alleviate this problem, recently researchers have strived to focus on Vision-Language Pre-training (VLP), which first~learns~contextualized multi-modal encoder representations over large-scale vision-language benchmarks, and then fine-tunes the pre-learnt multi-modal encoder on vision-language downstream tasks.

The main inspiration of recent attempts on VLP task \citep{chen2019uniter,huang2021multilingual,li2021scheduled,li2019unicoder,li2019visualbert,lu2019vilbert,luo2021video,pan2020auto,su2019vl,sun2019videobert,tan2019lxmert} are from the advances in natural language pre-training (e.g., BERT \citep{devlin2019bert}). The main idea of BERT is to derive bidirectional encoder representations from a Transformer based language model pre-trained on a large monolingual corpus. Two proxy tasks, i.e., masked word prediction and next sentence prediction, are designed to learn generic contextualized word representations, which facilitate multiple language understanding downstream tasks.
Following this philosophy, it is natural to pre-train a Transformer based structure with the multi-modal inputs (e.g., image-sentence pairs) instead of the monolingual sentences for vision-language understanding, which acts as multi-modal encoder to produce joint representations of visual content and natural sentence. Such pre-learnt multi-modal encoder enables the multi-modal reasoning and thus can be naturally adapted to vision-language perception tasks (e.g., visual question answering).

While impressive performances are reported, most existing BERT-like VLP approaches are not applicable to vision-language generation tasks (e.g., image captioning), which needs an additional language decoder for sentence generation conditioned on multi-modal inputs. It is not feasible to directly apply such BERT-like VLP methods to vision-language generation, since the sole pre-training process of encoder ignores the synchronous learning of a coupled decoder for generic language modeling. This inevitably results in the discrepancy between the pre-trained multi-modal encoder and the uninitialized language decoder that needs a separate initialization at downstream task, which might hinder the generalization of pre-training.
Therefore, how to simultaneously pre-train encoder for multi-modal representation extraction and language decoder for sentence generation is of great potential and importance~for VLP.

\sloppy{}

In this paper, we propose to mitigate this issue by pre-training a \textbf{Uni}versal \textbf{E}ncoder-\textbf{DE}coder \textbf{N}etwork (\textbf{Uni-EDEN}) to support both vision-language perception and generation tasks. Uni-EDEN adopts a Transformer based encoder-decoder structure, which consists of object and sentence encoders to separately encode the inputs of each modality (the set of detected image regions/word sequence), coupled with a sentence decoder that exploits the inter-modal interaction in between for both multi-modal reasoning and sentence generation. Moreover, as shown in Figure \ref{fig:figintro}, the linguistic representations of visual content (image region/entire image) can span different granularities including, from simple to comprehensive, individual label, a phrase, and a natural sentence.
To better align the visual content to linguistic representations in these different granularities, vision-language pre-training should take the hierarchical multi-granularity into consideration, which in turn endows the pre-learnt encoder-decoder with the power of multi-granular vision-language grounding.
Therefore, we present a novel array of multi-granular vision-language proxy tasks to pre-train Uni-EDEN.
Specifically, each vision-language granularity is modeled as a proxy task: 1) \emph{masked object classification}, the encoder takes a set of image regions with masked one as inputs and the decoder predicts the object reflected in this masked region conditioned on multi-modal encoder representations; 2) \emph{masked phrase generation}, the decoder should reconstruct the phrase that depicts the image region which is masked on the encoder side; 3) \emph{image-sentence matching}, the decoder discriminates whether the given image and sentence correspond to each other, and thus enforces the vision-language alignment at sentence level; 4) \emph{masked sentence generation}, to mimic the process of sentence generation, the decoder auto-regressively reconstructs each word of the input sentence depending on the input unmasked image regions and all ``past" words by preventing the ``future" words to be attended with an attention mask. In this sense, by equipping pre-training with multi-granular vision-language proxy tasks, all encoders and decoder in Uni-EDEN are simultaneously adapted to develop the capability of multi-modal representation extraction and language modeling.

To summarize, the main contributions of this work include: 1) we propose Uni-EDEN, a pre-trainable universal encoder-decoder network that facilitates both vision-language perception and generation tasks; 2) the design of multi-granular vision-language proxy tasks enables the alignment of visual content and linguistic representations at different granularities; 3) we adapt pre-learnt Uni-EDEN to a variety of downstream tasks (visual question answering, caption-based image retrieval, visual commonsense reasoning, and image captioning) and achieve new state-of-the-art performance for each task, which consistently validates the generalizability of Uni-EDEN.

\begin{figure*}[!tb]
\vspace{-0.1in}
\centering {\includegraphics[width=1\textwidth]{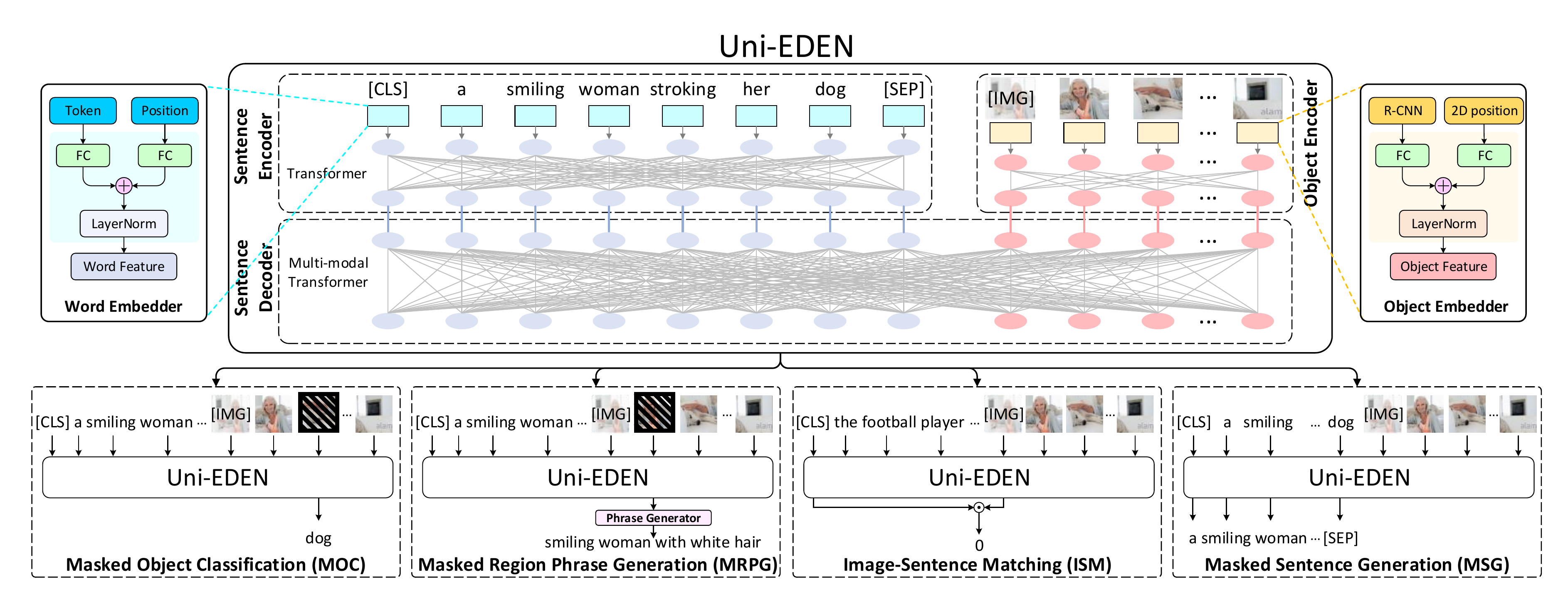}}
\vspace{-0.2in}
\caption{\small An overview of our Uni-EDEN, which is a two-stream Transformer based structure consisting of three module: object and sentence encoders that separately learns the representations of input region/word tokens, and sentence decoder that captures inter-modal interaction for both multi-modal reasoning and sentence generation. We pre-train Uni-EDEN with a series of multi-granular vision-language proxy tasks (MOC, MRPG, ISM, and MSG), aiming to align visual content to linguistic representations in different granularities, from individual label, phrase, to natural sentence.}
\label{fig:figframework}
\vspace{-0.1in}
\end{figure*}

\section{Related Work}
\noindent\textbf{Natural Language Pre-training.}
Recently, research on natural language pre-training has attracted increasing attention, and obtained impressive performances in multiple language understanding tasks.
One of the early successes for natural language pre-training is GPT \citep{radford2018improving} that pre-trains a Transformer based language model to extract general language representations depending on unidirectional word context. ELMo \citep{peters2018deep} includes contextualized word representations from pre-trained language models as additional feature to enhance language understanding downstream tasks.
BERT \citep{devlin2019bert} further upgrades GPT by involving masked language modeling proxy tasks during pre-training, which enables the learning of bidirectional representations conditioned on both left and right word contexts. Unlike BERT that is only applicable to language understanding via one encoder, MASS \citep{song2019mass} pre-trains an encoder-decoder model for language generation via masked sequence to sequence learning proxy tasks. Mostly recently, BART \citep{lewis2019bart} generalizes BERT for both language understanding and generation by combining bidirectional and auto-regressive
transformers for pre-training. Taking the inspiration from MASS and BART, our work pursuits their vision-language counterpart by pre-training a universal encoder-decoder structure and fine-tuning it to both vision-language perception and generation tasks.

\noindent\textbf{Vision-Language Pre-training.}
Sparked by natural language pre-training, a new wave of vision-language pre-training methods have been proposed recently to learn pre-trainable multi-modal encoders for vision-language perception tasks.
VisualBERT \citep{li2019visualbert} directly extends BERT by pre-training a Transformer based encoder with two visually-grounded language model objectives: masked language modeling with the image and  image-sentence matching. UNITER \citep{chen2019uniter}, Unicoder-VL \citep{li2019unicoder}, and VL-BERT \citep{su2019vl} further introduce masked region modeling proxy tasks to enhance the vision-language alignment during pre-training.
Unlike the aforementioned approaches that exploit a single-stream Transformer with multi-modal inputs for vision-language pre-training, ViLBERT \citep{lu2019vilbert} and LXMERT \citep{tan2019lxmert} present a two-stream Transformer based encoder structure. The input image regions and sentences are independently processed through two separate encoders, the outputs of which are further fused with a cross-modal encoder.

Comparatively few existing works have considered applying pre-training techniques for vision-language generation task. VideoBERT \citep{sun2019videobert} adapts BERT to learn multi-modal representations in video domain and addresses both action classification and video captioning tasks.
In this work, rather than pre-training multi-modal encoder solely in VideoBERT, we simultaneously pre-train the encoders and decoder in Uni-EDEN. Such design fully exploits the~merit of pre-training and endows Uni-EDEN with the capabilities of both multi-modal reasoning and language modeling.
Unified VLP \citep{zhou2019unified} is perhaps the most related~work, which learns a pre-trainable single-stream Transformer based encoder-decoder and also targets for both vision-language perception and generation.
Compared to Unified VLP, our Uni-EDEN differs in multiple ways: we utilize a more detailed two-stream design for Transformer based encoder-decoder structure, and we utilize a novel array of multi-granular vision-language proxy tasks to better align visual content to linguistic representations in different granularities, from individual label, phrase, to natural~sentence.

\section{Approach}

In this paper, we design a pre-trainable Universal Encoder-DEcoder Network (Uni-EDEN) that facilitates both vision-language perception and generation tasks.
The overall architecture of our Uni-EDEN is illustrated in Figure \ref{fig:figframework}.
We begin this section by elaborating the notation of VLP, followed with the detailed depiction of three components (i.e., object encoder, sentence encoder, and sentence decoder) in Uni-EDEN.
After that, four multi-granular vision-language proxy tasks are introduced to equip our Uni-EDEN with the abilities of both~multi-modal reasoning and language modeling conditioned on multi-granular vision-language grounding.
Finally, the overall objective of vision-language pre-training are presented.

\subsection{Notations}

In vision-language pre-training task, we are given a set of image-sentence pairs $\{\mathcal{I}, \mathcal{S}\}$ from a large-scale image captioning benchmark (Conceptual Captions \citep{sharma2018conceptual}).
The goal of this task is to pre-train an encoder-decoder network over the paired image-sentence data to extract multi-modal representations and auto-regressively generate sentence. The pre-trained encoder-decoder network can be further fine-tuned to support vision-language downstream tasks.

For each input image $\mathcal{I}$, we leverage the most common object detector (Faster R-CNN \citep{ren2015faster}) to detect objects within it, targeting for generalizing $\mathcal{I}$ into a set of salient image regions $\mathcal{R}_I = \left\{ \bm{r}_{i} \right\}^{N_I}_{i=1}$ containing $N_I$ detected objects. $\bm{r}_i \in \mathbb{R}^{D_r}$ denotes the $D_r$-dimensional visual representation of the $i$-th image region. Moreover, we denote the 2D position information of detected objects as $\mathcal{P}_I = \left\{ \bm{p}_{i} \right\}^{N_I}_{i=1}$. Here $\bm{p}_i \in \mathbb{R}^{D_p}$ ($D_p=5$) represents the geometric information of each bounding box consisting of the normalized top-left and bottom-right coordinates plus the proportion of image area covered.
For each corresponding sentence $\mathcal{S}$, we tokenize all words and represent it as a sequence of word tokens $\mathcal{W}_{S} = \left\{ \bm{w}_{j} \right\}^{N_S}_{j=1}$, where $\bm{w}_{j}\in \mathbb{R}^{D_w}$ is the one-hot encoding of $j$-th word token and $N_S$ is sentence~length.

\subsection{Model Architecture}

Inspired by recent progress in designing natural language pre-training models (e.g., Transformer in BERT), we construct the encoder and decoder modules in our Uni-EDEN with self-attention and multi-modal attention mechanisms \citep{vaswani2017attention}. Specifically, the whole model architecture takes the paired image regions and word token sequence as inputs and process them in a two-stream manner. Firstly, object and sentence encoders separately enhances the representations of each modality by introducing the position information of each region/word token.
The object/sentence encoder additionally refines each region/word token representation through self-attention mechanism that models intra-modal interactions.
After that, a following sentence decoder further strengthens the representations of each modality via multi-modal attention mechanism that induces the inter-modal interaction. Meanwhile, the sentence decoder learns to auto-regressively generate each word conditioned on the multi-modal representations, mimicking the process of sentence generation in image captioning task.

\textbf{Object Encoder.}
Object encoder operates over the input set of image regions and transforms them into a series of intermediate states, which are enhanced with the 2D position information of each region and the intra-modal contextual information.
Formally, we represent each image region (object) as a position-aware feature $\hat{\bm{r}}_{i}$ by fusing the visual feature $\bm{r}_i$ and the corresponding 2D position feature $\bm{p}_{i}$:
\begin{eqnarray}\label{Eq:Eq1}
\hat{\bm{r}}_{i}=\text{LayerNorm}(\bm{W}_{r} \bm{r}_i+\bm{W}_{p} \bm{p}_{i}),
\end{eqnarray}
where $\bm{W}_{r} \in \mathbb{R}^{D_H \times D_r}$ and $\bm{W}_{p} \in \mathbb{R}^{D_H \times D_p}$ are embedding matrices.
Note that we additionally include a special object token \texttt{[IMG]} that indicates the beginning of image region sequence, and its representation ${\bm{r}}_0$ is measured as the mean-pooled object representation ({\small ${\bm{r}}_0 = \frac{1}{N_I}\sum\nolimits_{i = 1}^{N_I} {{\bm{r}}_{i}}$}).
Next, the whole image region sequence $\hat{\mathcal{R}}_I = \left\{ \hat{\bm{r}}_{i}\right\}^{N_I}_{i=0}$ is fed into object encoder, which is implemented as $F$ stacked Transformer layers \citep{vaswani2017attention}.
Let ${\mathcal{H}}^{(f)}_I = \left\{ {\bm{h}}^{(f)}_{I,i}\right\}^{N_I}_{i=0}$ denote the output intermediate object representations of $f$-th Transformer layer.
Specifically, for each Transformer layer, it takes the outputs of previous Transformer layer $\hat{\mathcal{H}}^{(f)}_I$ as intra-modal context information to enhance each input object feature ${\bm{h}}^{(f)}_{I,i}$ via self-attention mechanism:
\begin{eqnarray}\label{Eq:Eq2}
{\bm{h}}^{(f+1)}_{I,i}=\texttt{Transformer}({\bm{h}}^{(f)}_{I,i},\hat{\mathcal{H}}^{(f)}_I).
\end{eqnarray}
Lastly, the object encoder outputs the final enhanced object representations ${\mathcal{H}}^{(F)}_I=\left\{ {\bm{h}}^{(F)}_{I,i}\right\}^{N_I}_{i=0}$, which reflect the intra-modal interactions among all objects.

\textbf{Sentence Encoder.}
Similarly, sentence encoder firstly enhances the input word representations by introducing the position of each word token.
The position-aware word representations are further strengthened with the intra-modal contextual information.
Concretely, given the input sequence of word tokens, we expand it by including two special word tokens \texttt{[CLS]} and \texttt{[SEP]} that indicate the beginning and ending of sentence. By denoting the one-hot encodings of \texttt{[CLS]} and \texttt{[SEP]} as $\bm{w}_{0}$ and $\bm{w}_{N_S+1}$, each word token is represented as a position-aware word feature $\hat{\bm{w}}_{j}$, which is measured as the combination of one-hot encoding $\bm{w}_{j}$ and the corresponding index embedding $\textbf{j}$:
\begin{eqnarray}\label{Eq:Eq3}
\hat{\bm{w}}_{j}=\text{LayerNorm}(\bm{W}_{w} \bm{w}_j+\bm{W}_{s} \textbf{j}),
\end{eqnarray}
where $\bm{W}_{w} \in \mathbb{R}^{D_H \times D_w}$ and $\bm{W}_{s} \in \mathbb{R}^{D_H \times D_j}$ are embedding matrices.
After that, the whole word token sequence $\hat{\mathcal{W}}_S = \left\{ \hat{\bm{w}}_{j}\right\}^{N_S+1}_{i=0}$ is fed into the $M$ stacked Transformer layers.
Here we denote the output intermediate word representations of $m$-th Transformer layer as ${\mathcal{H}}^{(m)}_S = \left\{ {\bm{h}}^{(m)}_{S,j}\right\}^{N_S+1}_{j=0}$.
In particular, each transformer layer performs self-attention over the outputs of previous Transformer layer $\hat{\mathcal{H}}^{(m)}_S$, and thus enhances each word feature ${\bm{h}}^{(m)}_{S,j}$ with intra-modal context information from $\hat{\mathcal{H}}^{(m)}_S$:
\begin{eqnarray}\label{Eq:Eq4}
{\bm{h}}^{(m+1)}_{S,j}=\texttt{Transformer}({\bm{h}}^{(m)}_{S,j},\hat{\mathcal{H}}^{(m)}_S).
\end{eqnarray}
Accordingly, the final output word features of sentence encoder is denoted as ${\mathcal{H}}^{(M)}_S=\left\{{\bm{h}}^{(M)}_{S,j}\right\}^{N_S+1}_{j=0}$, which reflect the intra-modal interactions in between.

\textbf{Sentence Decoder.}
The sentence decoder collects the refined object and word token features (${\mathcal{H}}^{(F)}_I$, ${\mathcal{H}}^{(M)}_S$) from each encoder, and simultaneously feeds them into a stack of $K$ multi-modal Transformer layers. Each multi-modal Transformer layer strengthens the representations of each modality by exploiting the inter-modal interaction through multi-modal attention mechanism.
Specifically, let ${\mathcal{H}}^{(k)}_{d_I} = \left\{ {\bm{h}}^{(k)}_{{d_I},i}\right\}^{N_I}_{i=0}$ and ${\mathcal{H}}^{(k)}_{d_S} = \left\{ {\bm{h}}^{(k)}_{{d_S},j}\right\}^{N_S+1}_{j=0}$ denote the output intermediate object and word features of $k$-th multi-modal Transformer layer.
For each multi-modal Transformer layer, it first concatenates the given object and word intermediate features as multi-modal input (${\mathcal{H}}^{(k)}_{d_{SI}}=[{\mathcal{H}}^{(k)}_{d_S},{\mathcal{H}}^{(k)}_{d_I}]$), and then feeds ${\mathcal{H}}^{(k)}_{d_{SI}}$ into transformer block. In this way, each region/word feature (${\bm{h}}^{(k)}_{{d_I},i}$/${\bm{h}}^{(k)}_{{d_S},j}$) is enhanced with inter-modal context information from ${\mathcal{H}}^{(k)}_{d_{SI}}$:
\begin{eqnarray}\label{Eq:Eq5}
{\bm{h}}^{(k+1)}_{{d_I},i}=\texttt{Transformer}({\bm{h}}^{(k)}_{{d_I},i},{\mathcal{H}}^{(k)}_{d_{SI}}),\\
{\bm{h}}^{(k+1)}_{{d_S},j}=\texttt{Transformer}({\bm{h}}^{(k)}_{{d_S},j},{\mathcal{H}}^{(k)}_{d_{SI}}).
\end{eqnarray}
As such, the final output multi-modal features are denoted as $\small{{\mathcal{H}}^{(K)}_{d_{SI}} = [{\mathcal{H}}^{(K)}_{d_S},{\mathcal{H}}^{(K)}_{d_I}]=\big[\left\{ {\bm{h}}^{(K)}_{{d_I},i}\right\}^{N_I}_{i=0},\left\{ {\bm{h}}^{(K)}_{{d_S},j}\right\}^{N_S+1}_{j=0}\big]}$ for multi-modal reasoning.
Moreover, conditioned on the multi-modal representation of each input word token ${\bm{h}}^{(K)}_{{d_S},j}$, the sentence decoder learns to auto-regressively predict the next word, leading to the output sentence word-by-word.
Please note that more details for sentence generation can be referred in masked sentence generation proxy task (Section \ref{MSG}).

\subsection{Multi-Granular Vision-Language Proxy Tasks}

An image is essentially a rich scenery that can be expressed with multi-granular linguistic representations.
For instance, the visual content in an image (e.g., image region or entire image) can be represented by a hierarchical linguistic structure including, from simple to comprehensive, individual label, a phrase, and a natural sentence.
Different granularity depicts distinct capability to understand visual content. Multi-modal reasoning in vision-language tasks should take this intrinsic linguistic structure into account.
Motivated by the above observations, we design a novel array of multi-granular vision-language proxy tasks to pre-train our Uni-EDEN by aligning visual content to linguistic representations in these different granularities.
In this sense,~the pre-trained Uni-EDEN is endowed with the power of multi-granular vision-language grounding, which facilitates multi-modal reasoning for both vision-language perception and generation.

\textbf{Masked Object Classification (MOC).}
In analogy to Masked Language Modeling (MLM) in BERT, we include the MOC proxy task that enforces the sentence decoder to reconstruct the object reflected in each masked image region conditioned on the unmasked regions and word tokens.
In this way, MOC drives our Uni-EDEN to not only capture the dependencies among image regions, but also ground each object label to the corresponding region.
Formally, we randomly mask the input image regions (15\% probability) and each masked region ${\bm{r}}^\mathbf{m}_i$ is replaced with a special token \texttt{[MASK]}. After feeding the word token sequence and image region sequence with masked ones $\{{\bm{r}}^\mathbf{m}_i\}^{\hat M}_{i=1}$ into Uni-EDEN, we can leverage the output multi-modal representation of each masked region ${\bm{r}}^\mathbf{m}_i$ to estimate the object label (i.e., object distribution $s_{\theta}({\bm{r}}^\mathbf{m}_i)$) through a softmax layer.
To reconstruct the object labels of masked regions, the objective of MOC is thus defined as the KL divergency loss to measure the mismatch between the estimated object distribution and ground-truth object~distribution:
\begin{eqnarray}
\mathcal{L}_{\text{MOC}}(\theta) = E_{(\mathcal{I}, \mathcal{S})\sim B} \sum_{i=1}^{\hat M} KL( g({\bm{r}}^\mathbf{m}_i) || s_{\theta}({\bm{r}}^\mathbf{m}_i) ),
\end{eqnarray}
where $B$ is the whole dataset. $g({\bm{r}}^\mathbf{m}_i)$ denotes the ground-truth object distribution of ${\bm{r}}^\mathbf{m}_i$, which is directly obtained from the off-the-shelf object detector used in region feature extraction.

\textbf{Masked Region Phrase Generation (MRPG).} Considering that image region usually conveys more semantics beyond the individual object label (e.g., adjectives of object), MOC that infers the general object label of masked region will fail to exploit all semantic cues in region for vision-language pre-training.
In MRPG, we go one step further and target at reconstructing a more comprehensive linguistic representation (i.e., phrase) of masked region.
In particular, given the output multi-modal feature of each masked region ${\bm{r}}^\mathbf{m}_i$, we feed it into a LSTM-based phrase generator to generate phrase word-by-word.
To supervise the process of phrase reconstruction, we adopt a phrase generator \citep{vinyals2015show} pre-trained on Visual Genome \citep{krishna2017visual} to produce the ground-truth phrase $\mathcal{P}({\bm{r}}^\mathbf{m}_i)$ for each masked region ${\bm{r}}^\mathbf{m}_i$.
The whole model is thus optimized by minimizing the negative log probability of the ground-truth phrase $\mathcal{P}({\bm{r}}^\mathbf{m}_i)$ given the unmasked image regions ${\bm{r}}^{\setminus \mathbf{m}}$ and the input sentence $\mathcal{S}$:
\begin{eqnarray}
\mathcal{L}_{\text{MRPG}}(\theta) = - E_{(\mathcal{I}, \mathcal{S})\sim B} \sum_{i=1}^{\hat M} \log Pr^p_{\theta}(\mathcal{P}({\bm{r}}^\mathbf{m}_i)|{\bm{r}}^{\setminus \mathbf{m}},\mathcal{S}).
\end{eqnarray}

\textbf{Image-Sentence Matching (ISM).}
Some vision-language perception tasks (e.g., visual question answering and caption-based image retrieval) capitalize on the understanding of holistic image-sentence relationship, which can not be directly captured via MOC and MRPG that only exploit the local contextual information. To enforce Uni-EDEN to understand the holistic multi-modal relationship, we additionally pre-train Uni-EDEN with ISM proxy task that pursues the vision-language alignment at sentence-level.
Specifically, given the inputs of image-sentence pair $\{\mathcal{I}, \mathcal{S}\}$, we feed the multi-modal outputs of object and sentence encoders into an attention-based two-layer MLP \citep{yu2019deep} to measure the image-sentence matching score $s_{\theta}(\mathcal{I}, \mathcal{S})$, that denotes how well the input sentence is semantically matched with image. In ISM, Uni-EDEN is optimized with triplet ranking loss, which enforces the matching score of matched pair $\{\mathcal{I}, \mathcal{S}\}$ to be larger than the score of mismatched pair $\{\mathcal{I}, \mathcal{S}^{-}\}$:
\begin{eqnarray}\label{Eq:Eq4}\small
\begin{aligned}
\mathcal{L}_{\text{ISM}}(\theta) =  E_{(\mathcal{I}, \mathcal{S})\sim B} [max(0, h- s_{\theta}(\mathcal{I}, \mathcal{S})
+ s_{\theta}(\mathcal{I}, \mathcal{S}^{-}))],
\end{aligned}
\end{eqnarray}
where $h$ represents the margin in the triplet ranking loss. Please note that the mismatched image-sentence pair is generated by replacing the image/sentence in a matched pair with a randomly sampled one from other pairs.

\textbf{Masked Sentence Generation (MSG).}\label{MSG}
All the aforementioned three vision-language proxy tasks focus on developing the capability of multi-modal reasoning, but fail to teach Uni-EDEN how to auto-regressively decode words for sentence generation. This severely hinders the generalization of pre-trained Uni-EDEN for vision-language generation downstream task (e.g., image captioning \cite{anderson2017bottom,li2019pointing,vinyals2015show} and video captioning \cite{chen2019temporal,li2018jointly,pan2016jointly}).
Hence, in order to mimic the process of sentence generation during pre-training, the MSG task is involved to enforce the sentence decoder to reconstruct the whole sentence word-by-word depending on the input image regions.
Concretely, at $j$-th decoding time-step, we aim to predict the $j$-th word $\bm{w}_{j}$ in sentence $\mathcal{S}$ conditioned on the input image $\mathcal{I}$ and all the ``past" words $\mathcal{S}_{0:j-1}$.
To implement this, we introduce an attention mask in Transformer and Multi-modal Transformer layers that prevents all the ``future" words $\mathcal{S}_{j:N_S}$ to be attended.
Thus, given the image region sequence and word token sequence with attention mask, the output multi-modal representation of $\bm{w}_{j-1}$ is utilized to predict the next word $\bm{w}_{j}$.
Finally, the objective of MSG can be expressed as the joint negative log probability for reconstructing the sequential words depending on all the ``past" words $\mathcal{S}_{0:j-1}$ and input image~$\mathcal{I}$:
\begin{eqnarray}
\mathcal{L}_{\text{MSG}}(\theta) = - E_{(\mathcal{I}, \mathcal{S})\sim B} \sum_{j=1}^{N_S+1} \log Pr_{\theta}(\bm{w}_{j}|\mathcal{S}_{0:j-1},\mathcal{I}).
\end{eqnarray}

\subsection{Connections between Multi-Granular Proxy Tasks and Downstream Tasks}

In this section, we conduct a detail discussion to explore the relationship between the multi-granular vision-language proxy tasks and downstream tasks. Specifically, the proxy task of masked object classification encourages the grounding of each object label to the corresponding region. Such capability of region-object alignment supports the fundamental region-level multi-modal reasoning, and thus facilitates most downstream tasks (e.g., VQA and VCR). The proxy task of masked phrase generation enforces Uni-EDEN to capture more semantics beyond the individual object label (e.g., adjectives of object). Such comprehensive semantics will not only boost the region-level multi-modal reasoning in VQA/VCR downstream tasks, but also enrich the generated captions for image captioning downstream task. The proxy task of image-sentence matching exploits the holistic image-sentence relations, and thus facilitates the caption-based image retrieval downstream task. Finally, the proxy task of masked sentence generation naturally mimics the process of sentence generation, and enables the generalization of pre-trained encoder-decoder structure for vision-language generation downstream tasks (e.g., image captioning).

\subsection{Overall Objective}

During pre-training, the overall training objective of Uni-EDEN integrates the loss functions of four multi-granular vision-language proxy tasks, and the masked language modeling objective ($\mathcal{L}_{\text{MLM}}(\theta)$) as in BERT:
\begin{eqnarray}
\begin{aligned}
\mathcal{L} = \mathcal{L}_{\text{MOC}}(\theta)+&\mathcal{L}_{\text{MRPG}}(\theta)+\mathcal{L}_{\text{ISM}}(\theta)+\mathcal{L}_{\text{MSG}}(\theta)+\mathcal{L}_{\text{MLM}}(\theta).
\end{aligned}
\end{eqnarray}

\section{Experiments}

We pre-train our Uni-EDEN on an automatically collected large-scale image captioning dataset (Conceptual Captions \citep{sharma2018conceptual}), and evaluate the generalization of pre-trained Uni-EDEN by fine-tuning it on four vision-language downstream tasks, i.e., visual question answering, caption-based image retrieval, visual commonsense reasoning, and image captioning.

\subsection{Pre-training Dataset and Settings}
The experiments for pre-training are conducted on the large-scale image captioning benchmark (Conceptual Captions \citep{sharma2018conceptual}). Conceptual Captions contains 3.3 million image-sentence pairs, which are automatically collected from billions of webpages. The caption of each image is programmatically created based on the original Alt-text from HTML webpages.

During pre-training, the adopted off-the-shelf Faster-RCNN is pre-trained on ImageNet \citep{ImageNet} and Visual Genome \citep{krishna2017visual} as in \citep{anderson2017bottom}.
At most 100 image regions with detection confidences higher than 0.2 are selected as inputs. Each input image region representation is a 2,048-dimensional vector.
For the proxy task of masked phrase generation, we leverage the LSTM-based phrase generator \citep{vinyals2015show} pre-trained on region-phrase pairs in Visual Genome \citep{krishna2017visual}.
In object/sentence encoder, we stack $F=6$/$M=12$ Transformer layers. The sentence decoder includes $K=6$ Multi-modal Transformer layers. The margin $h$ is set as 0.2 as in \citep{lee2018stacked}.
The whole vision-language pre-training architecture are mainly implemented with PyTorch \citep{paszke2019pytorch}, optimized with Adam \citep{kingma2014adam} on 16 Tesla V100 GPUs.
We set the mini-batch size as 512 and the learning rate as 0.0001. The maximum iteration is 10~epoches.

\begin{figure*}[!tb]
\vspace{-0.0in}
\centering {\includegraphics[width=1\textwidth]{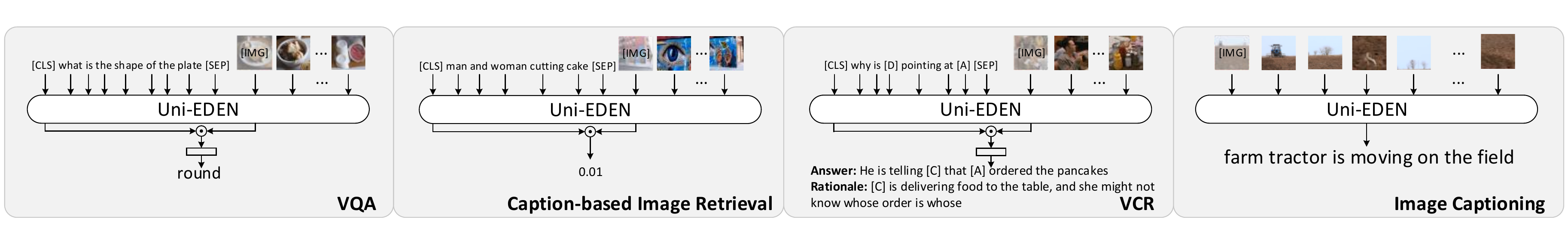}}
\vspace{-0.26in}
\caption{\small The fine-tuning procedures on four vision-language downstream tasks.}
\label{fig:figdownstream}
\vspace{-0.1in}
\end{figure*}

\subsection{Fine-Tuning on Downstream Tasks}
This section details the fine-tuning procedure of each vision-language downstream task (as shown in Figure \ref{fig:figdownstream}), coupled with the experimental settings.

\noindent\textbf{Visual Question Answering (VQA).}
In VQA, the model predicts an answer to the given natural language question with regard to an image.
In this task, we adopt the VQA 2.0 dataset \citep{antol2015vqa} for fine-tuning our Uni-EDEN, which contains 1.1 million questions about images in COCO \citep{chen2015microsoft}.
During fine-tuning, we follow the official split \citep{anderson2017bottom} and frame this task as a multi-label classification problem.
In particular, by feeding the input image-question pair into Uni-EDEN, we learn the holistic image-question representation by fusing output multi-modal features of object and sentence encoders via attention mechanism \citep{yu2019deep}.
A single-layer MLP is further leveraged to predict answer by embedding the holistic image-question representation into 3,129 possible answers.
At fine-tuning stage, as in \citep{anderson2017bottom,zhou2019unified}, we optimize the output answer predictions in Uni-EDEN with regard to the soft answer labels via cross-entropy loss.
The mini-batch size is 96 and the learning rate is 0.00005. We stop the fine-tuning procedure after 20~epoches.

\noindent\textbf{Caption-based Image Retrieval (CIR).}
CIR aims to search an image from a pool given its caption that depicts the image content.
The dataset adopted in this task is Flickr30k \citep{plummer2015flickr30k} consisting of 31k images from Flickr.
Each image is equipped with five human-annotated sentences.
We utilize the commonly adopted split in \citep{lee2018stacked} and formulate this task as a ranking problem that sorts images according to the learnt image-sentence matching scores.
Specifically, for each matched image-sentence pair, we construct one mismatched pair by hard example mining.
During fine-tuning, we feed each matched/mismatched pair into Uni-EDEN and obtain the matching score $s_{\theta}(\mathcal{I}, \mathcal{S})$ as in image-sentence matching pre-training task.
The whole model is thus optimized with triplet ranking loss, that aims to enforce the matching score of matched pair to be larger than the score of mismatched pair.
We set the mini-batch size as 512 and the learning rate as 0.00002. The maximum iteration is set as 30 epoches.

\noindent\textbf{Visual Commonsense Reasoning (VCR).}
VCR tackles two problems: visual question answering (Q$\to$A) and answer justification (QA$\to$R), that requires the model to predict an answer or judge the correctness of the chosen rationale respectively. Each problem is framed as multiple choice task. In addition, VCR includes a holistic setting (Q$\to$AR) that the model should choose the right answer (from four answer choices) and then select the correct rationale for that answer (from four rationale choices).
The Visual Commonsense Reasoning (VCR) benchmark \citep{zellers2019recognition} is utilized for evaluation in this task.
The dataset includes 290k multiple choice QA problems from 110k movie scenes.
During fine-tuning, we concatenate the question and each possible response (answer or rationale) as the textual input, which is fed into Uni-EDEN along with the image.
As in VQA, we measure the holistic image-sentence feature based on the multi-modal outputs of object and sentence encoders, and then utilize a linear layer to predict the score for each possible response.
The final prediction (i.e., all scores of the four response choices) is thus trained with cross-entropy loss (mini-batch size: 64, learning rate: 0.00002, maximum iteration: 20 epoches).

\noindent\textbf{Image Captioning (IC).}
In IC, the model aims to auto-regressively generate the natural sentence that depicts the visual content of input image.
We use COCO \citep{chen2015microsoft}, the most popular image captioning benchmark, for fine-tuning and evaluating our Uni-EDEN.
COCO contains 123,287 images, and each image is annotated with five sentences.
Here we utilize the widely adopted Karpathy split \citep{Karpathy:CVPR15,yao2019hierarchy,yao2017boosting,pan2020x} for evaluation, consisting of 113,287 images for training, 5k images for validation, and 5k images for testing.
For fine-tuning, we firstly optimize the whole architecture with cross-entropy loss. The mini-batch size is 16 and the learning rate is 0.00003. We set the maximum iteration as 10 epoches.
The fine-tuned Uni-EDEN is further trained with self-critical training strategy \citep{rennie2017self}, which enables sequence-level optimization with CIDEr reward. The learning rate is 0.000005 and the maximum iteration is 30 epoches.

\begin{table*}[!tb]\scriptsize
\vspace{-0.1in}
\centering
\setlength{\extrarowheight}{2.0pt}
\setlength\tabcolsep{1.1pt}
\caption{Performance comparison with state-of-the-art (\textbf{SOTA}) task-specific models and vision-language pre-training techniques (\textbf{Pre-E}: pre-trainable encoder module; \textbf{Pre-ED}: pre-trainable encoder-decoder structure) on four vision-language downstream tasks. Uni-EDEN$^-$ is one variant of our Uni-EDEN which is directly trained with task-specific training data, without pre-training. For VQA which has private test sets, we report test results (in parentheses) for our Uni-EDEN. $\star$ denotes our implementation by using the same pre-training data/backbone as in Uni-EDEN.}
\vspace{-0.1in}
\label{table:exp}
\begin{tabular}{c|l|cccc|cccc|ccc|ccc}
   \Xhline{2\arrayrulewidth}
   & \multirow{2}{*}{Model} & \multicolumn{4}{c|}{Image Captioning}  & \multicolumn{4}{c|}{VQA} & \multicolumn{3}{c|}{CIR}  & \multicolumn{3}{c}{VCR} \\
   & ~ &  B@4  &  M  &  C  &  S  &  Overall  &  Yes/No  &  Number  &  Other  &  R1  &  R5  &  R10  &  Q $\rightarrow$ A  &  QA $\rightarrow$ R  &  Q $\rightarrow$ AR  \\ \hline\hline

\multirow{7}{*}{\textbf{SOTA}}
        & BUTD \citep{anderson2017bottom}     & 36.3 & 27.7 & 120.1 & 21.4 & 65.3 (65.7) & 81.8 & 44.2 & 56.1 & -    & -    & -    & -    & -    & -     \\
        & AoANet \citep{huang2019attention}      & 38.9 & 29.2 & 129.8 & 22.4 & -    & -    & -    & -    & -    & -    & -    & -    & -    & -     \\
        & BAN \citep{kim2018bilinear}         & -    & -    & -     & -    & 70.0 (70.4) & 85.4 & 54.0 & 60.5 & -    & -    & -    & -    & -    & -     \\
        & DFAF \citep{gao2019dynamic}        & -    & -    & -     & -    & 70.2 (70.3) & 86.1 & 53.3 & 60.5 & -    & -    & -    & -    & -    & -     \\
        & SCAN \citep{lee2018stacked}        & -    & -    & -     & -    & -    & -    & -    & -    & 48.6 & 77.7 & 85.2 & -    & -    & -     \\
        & R2C \citep{zellers2019recognition} & -    & -    & -     & -    & -    & -    & -    & -    & -    & -    & -    & 63.8 & 67.2 & 43.1  \\ \hline
        & Uni-EDEN$^-$ (single-stream) & 38.2 & 28.6 & 126.7 & 22.3 & 68.4 & 85.2 & 50.2 & 58.1 & 54.6 & 81.4 & 88.0 & 72.5 & 73.9 & 53.6  \\
		& Uni-EDEN$^-$ & 38.8 & 28.8 & 128.2 & 22.8 & 69.8 & 86.3 & 49.5 & 60.4 & 55.6 & 81.8 & 88.5 & 73.0 & 74.4 & 54.5  \\ \hline
\multirow{5}{*}{\textbf{Pre-E}}
        & VisualBERT \citep{li2019visualbert}  & -    & -    & -     & -    & 70.8 (71.0) & -    & -    & -    & -    & -    & -    & 70.8 & 73.2 & 52.2  \\
		& ViLBERT \citep{lu2019vilbert}     & -    & -    & -     & -    & 70.6 (70.9) & -    & -    & -    & 58.2 & 84.9 & 91.5 & 72.4 & 74.5 & 54.0  \\
        & VL-BERT \citep{su2019vl}     & -    & -    & -     & -    & 71.2 & -    & -    & -    & -    & -    & -    & 73.8 & 74.4 & 55.2  \\
        & LXMERT$^\star$ \citep{tan2019lxmert} & - & - & - & - & 70.6 & 87.2 & 52.6 & 60.4 & 61.0 & 86.8 & 92.1 & 73.2 & 73.8 & 54.4 \\
        & UNITER$^\star$ \citep{chen2019uniter} & - & - & - & - & 70.7 & 87.3 & 54.4 & 60.2 & 61.5 & 87.2 & 92.2 & 73.2 & 75.0 & 55.3 \\
        \hline
\multirow{2}{*}{\textbf{Pre-ED}}
        & Unified VLP (ResNet-101)$^\star$ \citep{zhou2019unified}  & 39.1 & 29.1 & 129.0 & 23.0 & 70.4 & 87.0 & 53.8 & 60.1 & 60.8  & 86.7 & 92.0 & 73.5 & 75.2 & 55.4     \\
        & Unified VLP (ResNeXt) \citep{zhou2019unified}  & 39.5 & 29.3 & 129.3 & 23.2 & 70.5 (70.7) & 87.2 & 52.1 & 60.3 & -    & -    & -    & -    & -    & -     \\
        \cline{2-16}
        & Uni-EDEN (single-stream)    & 39.7 & 29.4 & 131.8 & 23.2 & 71.3 & 87.5 & 54.0 & 61.3 & 62.1 & 87.1 & 92.2 & 74.2 & 76.1 & 56.8  \\
        & Uni-EDEN     & \textbf{39.9} & \textbf{29.6} & \textbf{133.2} & \textbf{23.4} & \textbf{72.2} (\textbf{72.5}) & \textbf{88.1} & \textbf{54.6} & \textbf{62.7} & \textbf{63.2} & \textbf{87.3} & \textbf{92.3} & \textbf{74.9} & \textbf{76.7} & \textbf{57.8}  \\
  \Xhline{2\arrayrulewidth}
\end{tabular}
\vspace{-0.0in}
\end{table*}

\subsection{Performance Comparison}
Table \ref{table:exp} summarizes the quantitative results of our Uni-EDEN on all the vision-language perception (VQA, CIR, and VCR) and generation (IC) downstream tasks.
We compare Uni-EDEN with state-of-the-art task-specific models and vision-language pre-training techniques in each vision-language downstream task.
Please note that we additionally include several variants of our Uni-EDEN: 1) Uni-EDEN$^-$ which is directly trained with task-specific training data for each task, without pre-training on Conceptual Captions dataset, and 2) Uni-EDEN (single-stream) that pre-trains the single-stream Transformer based encoder-decoder structure (as in Unified VLP \citep{zhou2019unified}) with our proposed multi-granular vision-language proxy tasks.
The SOTA task-specific models include BUTD \citep{anderson2017bottom} and AoANet \citep{huang2019attention} for IC, BUTD \citep{anderson2017bottom}, BAN \citep{kim2018bilinear} and DFAF \citep{gao2019dynamic} for VQA, SCAN \citep{lee2018stacked} for CIR, and R2C \citep{zellers2019recognition} for VCR.
The vision-language pre-training baselines are grouped into two directions: pre-trainable encoder module (i.e., VisualBERT \citep{li2019visualbert}, ViLBERT \citep{lu2019vilbert}, VL-BERT \citep{su2019vl}, LXMERT \citep{tan2019lxmert}, and UNITER \citep{chen2019uniter}) for only vision-language perception tasks, and pre-trainable encoder-decoder structure (Unified VLP \citep{zhou2019unified}) for both vision-language perception and generation tasks. For fair comparison with our Uni-EDEN, we re-implement LXMERT and UNITER by pre-training them over Conceptual Captions. We also include an additional run of Unified VLP, named as Unified VLP (ResNet-101), by using the same backbone (ResNet-101) as in our Uni-EDEN.

\noindent\textbf{Comparison with SOTA Task-specific Models.}
In general, under the same task-specific training setting without vision-language pre-training, our Uni-EDEN$^-$ achieves comparable results with other SOTA baselines on all vision-language tasks.
The results basically demonstrate the effectiveness of the adopted two-stream Transformer based structure with self-attention and multi-modal attention.
Furthermore, when pre-training Uni-EDEN on Conceptual Captions and then fine-tuning it on task-specific data, our Uni-EDEN consistently exhibits better performances than other SOTA task-specific baselines across all the evaluation metrics on four tasks.
Remarkably, for vision-language generation task (image captioning), Uni-EDEN can achieve 133.2\% with CIDEr score optimization, which is to-date the best performance without any model ensemble and makes the absolute improvement over the best competitor AoANet by 3.4\%. In addition, for vision-language perception task (VQA), Uni-EDEN outperforms DFAF by a large margin (2.0\% for Overall accuracy).
The performance improvements generally show the key advantage of exploiting vision-language pre-training via Uni-EDEN, that facilitate both vision-language perception and generation tasks.

To fully verify the generalizability of our proposed multi-granular vision-language proxy tasks, we include the variants of Uni-EDEN, i.e., Uni-EDEN (single-stream), that adopt the single-stream Transformer based backbone for VLP. In particular, our Uni-EDEN (single-stream) with additional vision-language pre-training outperforms Uni-EDEN$^-$ (single-stream) without pre-training on each downstream task. This again demonstrates the effectiveness of our multi-granular vision-language proxy tasks when equipped with a different single-stream Transformer based backbone.

\noindent\textbf{Comparison with VLP Techniques.}
Overall, the results across all vision-language perception and generation tasks consistently indicate that our Uni-EDEN obtains better performances against other state-of-the-art pre-trainable encoder modules (i.e., VisualBERT, ViLBERT, VL-BERT, LXMERT, and UNITER) and encoder-decoder structure (i.e., Unified VLP (ResNet-101) and Unified VLP (ResNeXt)).
In the IC and VQA tasks, the CIDEr and Overall of Uni-EDEN can achieve 133.2\% and 72.2\%, making 3.9\% and 1.0\% absolute improvements over the best competitors Unified VLP (ResNeXt) and VL-BERT, respectively.
In particular, by capitalizing on a multi-modal encoder that pre-learns multi-modal reasoning through vision-language pre-training, all the pre-trainable encoder modules lead to a large performance boost over SOTA task-specific approaches for vision-language perception tasks.
Nevertheless, the pre-trainable encoder modules can not be directly adapted to vision-language generation task (IC), which needs an additional language decoder for sentence generation.
By enabling the simultaneous pre-training of encoder and decoder, Unified VLP (ResNeXt) outperforms SOTA task-specific approaches in both vision-language perception and generation tasks.
Furthermore, by pre-training encoder-decoder structure with multi-granular vision-language proxy tasks, our Uni-EDEN boosts the performances in all the vision-language downstream tasks.
This clearly confirms the effectiveness of multi-granular vision-language proxy tasks that better align the visual content to linguistic representations in different granularities, from individual label, phrase, to natural sentence.

\begin{table*}[!tb]\small
\vspace{-0.1in}
\centering
\setlength{\extrarowheight}{2.0pt}
\setlength\tabcolsep{0.6pt}
\caption{\small Ablation study on the use of different vision-language proxy tasks for pre-training. \textbf{Base}: a base pre-training strategy by integrating masked language modeling, masked object classification, and image-sentence matching proxy tasks; \textbf{MSG}: Masked Sentence Generation task; \textbf{MRPG}: Masked Region Phrase Generation task.}
\vspace{-0.1in}
\label{table:ablation}
\begin{tabular}{ccc|cccc|cccc|ccc|ccc}
\Xhline{2\arrayrulewidth}
\multirow{2}{*}{\textbf{Base}} & \multirow{2}{*}{\textbf{MSG}} & \multirow{2}{*}{\textbf{MRPG}} & \multicolumn{4}{c|}{Image Captioning} & \multicolumn{4}{c|}{VQA} & \multicolumn{3}{c|}{CIR} & \multicolumn{3}{c}{VCR} \\
 &  &  & B@4 & M & C & S & Overall & Yes/No & Number  & Other & R1 & R5 & R10 & Q $\rightarrow$ A & QA $\rightarrow$ R & Q $\rightarrow$ AR \\ \hline\hline
$\checkmark$ &              &               & 39.2 & 28.9 & 129.1 & 22.8 & 71.5 & 87.7 & 53.0 & 61.8 & 61.8 & 87.1 & 92.2 & 74.6 & 75.9 & 56.9  \\
$\checkmark$ & $\checkmark$ &               & 39.7 & 29.4 & 132.4 & 23.2 & 71.8 & 88.0 & 53.6 & 62.2 & 62.5 & 86.7 & \textbf{92.3} & 74.8 & 76.1  & 57.3  \\
$\checkmark$ &              & $\checkmark$  & 39.3 & 29.2 & 130.6 & 22.9 & 72.0 & 88.0 & 54.1 & 62.5 & 62.2 & 86.7 & 92.2 & 75.0 & 76.2 & 57.4  \\
$\checkmark$ & $\checkmark$ & $\checkmark$  & \textbf{39.9} & \textbf{29.6} & \textbf{133.2} & \textbf{23.4} & \textbf{72.2} & \textbf{88.1} & \textbf{54.6} & \textbf{62.7} & \textbf{63.2} & \textbf{87.3} & \textbf{92.3} & \textbf{74.9} & \textbf{76.7} & \textbf{57.8} \\
  \Xhline{2\arrayrulewidth}
\end{tabular}
\vspace{-0.1in}
\end{table*}

\subsection{Ablation Study}

Here we conduct ablation study to investigate how each design of vision-language proxy task in our Uni-EDEN influences the performances of downstream tasks. Table \ref{table:ablation} details the results on four vision-language tasks by considering one more proxy task for pre-training Uni-EDEN.
We start from a base pre-training strategy which is a degraded version of our multi-granular vision-language proxy tasks by integrating the three commonly adopted ones (MLM, MOC, ISM).
This ablated base strategy solely targets for aligning visual contents to each individual label and holistic sentence for multi-modal reasoning, and obtains similar results to ViLBERT and VL-BERT on vision-language perception tasks.
After that, we extend the base strategy by including MSG proxy task, that aims to enforce Uni-EDEN to auto-regressively reconstruct input sentence.
This way naturally mimics the process of sentence generation during pre-training, thus significantly improves the generalization of Uni-EDEN on image captioning task.
Meanwhile, the base strategy can be upgraded by additionally involving the proxy task of MPRG that reconstructs a more comprehensive linguistic representation (phrase) of masked region. In this way, Base+MPRG achieves higher performances across all downstream tasks than Base. Furthermore, after integrating base strategy with both MSG and MPRG, consistent performance boosts are attained over both vision-language perception and generation tasks.
The results again demonstrate the merit of simultaneously exploiting the vision-language alignments at different granularities (i.e., object-label, object-phrase, and image-sentence alignments) for pre-training.

\begin{figure*}[!tb]
\vspace{-0.0in}
\centering {\includegraphics[width=0.95\textwidth]{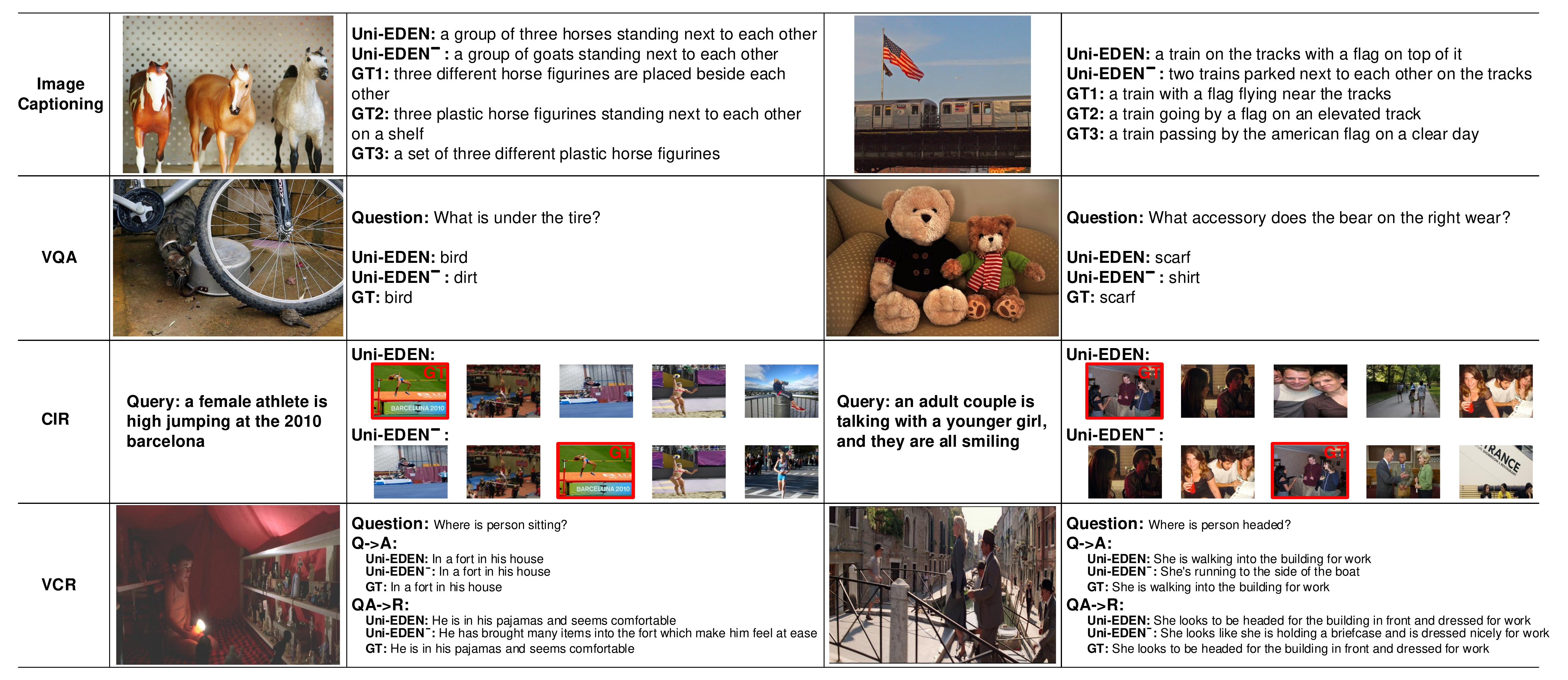}}
\vspace{-0.2in}
\caption{\small Qualitative Results on four vision-language downstream tasks. The output results (e.g., captions, answers or retrieval results) are generated by 1) Uni-EDEN, 2) Uni-EDEN$^-$, and 3) Ground Truth (GT).}
\label{fig:results}
\vspace{-0.1in}
\end{figure*}

\subsection{Qualitative Analysis}

\textbf{Qualitative Results.} Here we show the qualitative results of our Uni-EDEN$^-$ (without vision-language pre-training) and Uni-EDEN (with vision-language pre-training) on each downstream task. As shown in Figure \ref{fig:results}, our Uni-EDEN produces better captions, answers, and retrieval results, than Uni-EDEN$^-$ by exploiting additional pre-training via our proposed multi-granular vision-language proxy tasks. For example, in the downstream task of image captioning, Uni-EDEN$^-$ generates the phrase of ``a group of goats'' that is inconsistent with the visual content for the first image, while ``a group of three horses'' in our Uni-EDEN depicts the visual content more precise.

\begin{figure*}[!tb]
\vspace{-0.0in}
\centering {\includegraphics[width=0.95\textwidth]{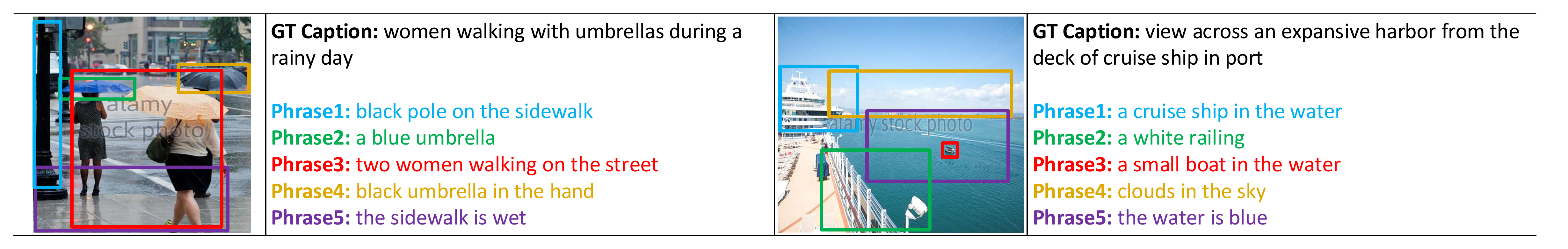}}
\vspace{-0.2in}
\caption{\small Visualization of region-phase alignment in our Uni-EDEN during pre-training.}
\label{fig:phrase}
\vspace{-0.2in}
\end{figure*}

\noindent\textbf{Visualization of Region-phase Alignment.} In order to better qualitatively evaluate the generated phrases of each region in our Uni-EDEN during pre-training, we visualize the region-phase alignment for image examples in Conceptual Captions dataset. As shown in Figure \ref{fig:phrase}, the phrase generator in Uni-EDEN can produce a more descriptive phrase for each region than the individual object label by enriching semantics with adjectives of object, thereby boosting vision-language pre-training with such region-phrase alignment. For example, compared to the individual object labels (e.g., ``pole'' and ``umbrella'') for the regions of the first image, the generated phrases (e.g., ``black pole on the sidewalk'' and ``a blue umbrella'') in our Uni-EDEN depicts the corresponding regions more comprehensive.

\section{Conclusions}
We have presented a pre-trainable Universal Encoder-DEcoder Network (Uni-EDEN) to facilitate both vision-language perception and generation tasks.
Particularly, we~study the problem from the viewpoint of simultaneously pre-training encoder for multi-modal representation extraction and decoder for language generation.
To materialize our idea, we construct a two-stream Transformer based~encoder-decoder structure that first learns the representations of each modality via object/sentence encoder, and further captures inter-modal interactions for multi-modal reasoning or sentence generation through sentence decoder.
Moreover, a novel array of multi-granular vision-language proxy tasks are presented to pre-train Uni-EDEN, targeting for better aligning the visual content to linguistic representations in different
granularities, from individual label, phrase, to natural sentence.
Extensive experiments demonstrate the compelling generalizability of Uni-EDEN by fine-tuning it to four vision-language tasks.
More remarkably, we obtain new state-of-the-art performance for each task with Uni-EDEN.

\textbf{Acknowledgments.}
This work was supported by the National Key R\&D Program of China under Grant No. 2020AAA0108600.

\bibliographystyle{ACM-Reference-Format}
\bibliography{egbib}

\end{document}